\documentclass[leqno, 10pt]{article}
\usepackage[margin=1in]{geometry}

\usepackage{url}
\usepackage{amsmath}
\usepackage{amsfonts}
\usepackage{graphicx}
\usepackage{array}
\usepackage{cite}
\usepackage{comment}
\usepackage{lmodern}
\usepackage{epstopdf}
\newtheorem{definition}{Definition}
\usepackage[T1]{fontenc}
\usepackage{textcomp}

\usepackage{times}  
\usepackage{helvet}  
\usepackage{courier}  

\usepackage{subcaption}

\usepackage{algorithmicx}
\usepackage{algcompatible}

\newtheorem{Theorem}{Theorem}[section]

\usepackage{float}
\floatstyle{ruled}
\newfloat{algorithm}{t}{lop}
\restylefloat*{algorithm}
\floatname{algorithm}{Algorithm}

\usepackage{etoolbox}

\usepackage[explicit]{titlesec}
\titlespacing\section{0pt}{2pt plus 4pt minus 0pt}{2pt plus 2pt minus 0pt}
\titlespacing\subsection{0pt}{2pt plus 4pt minus 0pt}{2pt plus 2pt minus 0pt}

\usepackage{enumitem}

\providecommand{\keywords}[1]{\textbf{\textit{Keywords: }} #1}

\newtheorem{defn}{Definition}

\begin{document}

\date{}
\title{Transfer Regression via Pairwise Similarity Regularization}

\author{Aubrey Gress\thanks{Department of Computer Science. University of California, Davis. agress@ucdavis.edu} \\
\and
Ian Davidson\thanks{Department of Computer Science. University of California, Davis. davidson@cs.ucdavis.edu}}

\maketitle

\begin{abstract}

Transfer learning methods address the situation where little labeled training data from the ``target'' problem exists, but much training data from a related ``source'' domain is available. However, the overwhelming majority of transfer learning methods are designed for simple settings where the source and target predictive functions are almost identical, limiting the applicability of transfer learning methods to real world data.  
We propose a novel, \textit{weaker}, property of the source domain that can be transferred even when the source and target predictive functions diverge.  Our method assumes the source and target functions share a ``Pairwise Similarity'' property, where if the source function makes similar predictions on a pair of instances, then so will the target function. We propose Pairwise Similarity Regularization Transfer, a flexible graph-based regularization framework which can incorporate this modeling assumption into standard supervised learning algorithms.  We show how users can encode domain knowledge into our regularizer in the form of spatial continuity, pairwise ``similarity constraints'' and how our method can be scaled to large data sets using the Nystr{\"o}m approximation.  Finally, we present positive and negative results on real and synthetic data sets and discuss when our Pairwise Similarity transfer assumption seems to hold in practice.

\end{abstract}

\keywords{Transfer Learning, Regression}

\section{Introduction}
\label{sec:introduction}

\begin{table*}[t]
\footnotesize
\begin{tabular}{| l | p{6cm} |  p{2.2cm} | p{4cm} |}
\hline
Type of Transfer & Transfer Assumption & Existing Work & Failure Condition \\
\hline
Covariate Shift& $P_T(X) \neq P_S(X)$, \newline $P_T(Y|X) = P_S(Y|X)$& \cite{sugiyama2007covariate, sugiyama2008direct, gretton2009covariate, pan2011domain} and many others& $P_T(Y|X) \neq P_S(Y|X)$ \\
\hline
Hypothesis Transfer & $f_S(x) \approx f_T(x)$ & \cite{tommasi2010safety, kuzborskij2013stability,patricia2014learning}  & $f_T(x)$ and $f_S(x)$ are too different \\
\hline
Location Scale & $P_T(Y|X) = \alpha(X)P_S(Y|X) + \delta(X)$ for smooth $\alpha(X)$ and $\delta(X)$& \cite{zhang2013domain, wang2014active, wang2014flexible}& $\alpha$  and $\delta$ are not smooth\\
\hline
Pairwise Similarity Transfer & If $f_S(x_i) \approx f_S(x_j)$ then $f_T(x_i) \approx f_T(x_j)$ & None & When this ``Pairwise Similarity'' assumption doesn't hold. \\
\hline
\end{tabular}
\caption{Popular settings addressed in the Transfer Learning literature, along with the assumptions they make, previous work and when these methods will fail.  Ours is the first to explore Pairwise Similarity Transfer. \label{tab:shift-assumptions}}
\end{table*}

\begin{figure}[t!]
  \centering
\includegraphics[width=.5\columnwidth]{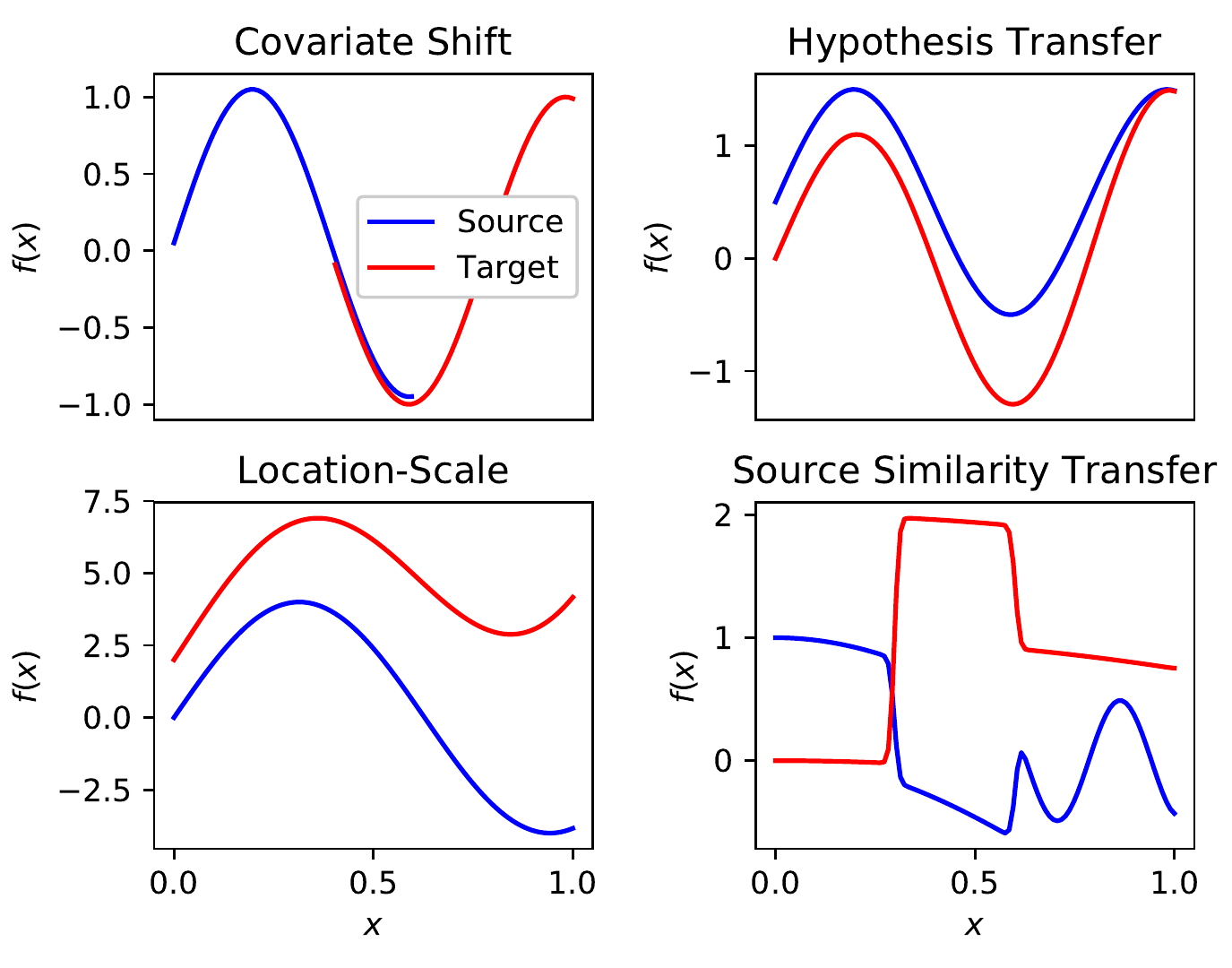}
  \caption{Examples of typical types of transfer learning.  Covariate Shift assumes the source and target posteriors are identical, but the densities $p(x)$ are different.  Hypothesis Transfer assumes the source and target functions are similar but not identical.  Location-Scale  allows different posteriors, but assumes a simple scaling and translation transformation connects them.  Pairwise Similarity Transfer explores a new setting, where if the source makes similar predictions on a pair of instances, then so will the target. 
  }
  \label{fig:example-shift}
\end{figure}

\noindent
\textbf{Motivation.} Standard supervised learning methods can require large amounts of labeled training data in order to learn an accurate function.  Transfer learning methods mitigate this by using a well labeled source task $S$ to aid the poorly labeled target task $T$.  While previous work has been successful in a broad range of machine learning settings, traditional transfer learning methods assume the source and target predictive functions $f_S(x)$ and $f_T(x)$ are very similar \cite{pan2010survey} (Figure \ref{fig:example-shift}, top-left and top-right).  As such, these methods will perform poorly if there are significant differences between these two functions.

As an example, consider the problem of predicting taxi pickups throughout San Francisco
given a source data set of neighborhood housing prices throughout the city.  While it's reasonable to assume there is some relationship between the two problems, such as pricier neighborhoods requiring fewer pickups due to car ownership, this relationship between the two predictive functions may be some non-trivial, non-constant and non-smooth transformation.  In this case, standard methods such as simply regularizing target estimates to be similar to source estimates may result in worse performance than just using the the small target data set, a phenomenon known as \textbf{negative transfer} \cite{pan2010survey}

\noindent
\textbf{Pairwise Similarity Transfer.} 

Consider the top-left, top-right, and bottom-left styles of transfer learning in Figure \ref{fig:example-shift}. Though mathematically different (see Table \ref{tab:shift-assumptions}), they all effectively transfer the \textbf{shape} of the source function when estimating the target function, essentially requiring the target and source functions to be identical or near-identical. However, in our example with home prices and taxi-pickups this is unlikely to be the case.  In this work we examine a new property that can be shared by the source and target, ``Pairwise Similarity,'' wherein if the source function makes similar predictions on a pair of instances (i.e. $f_S(x_i) \approx f_S(x_j)$) then so will the target function (i.e. $f_T(x_i) \approx f_S(x_j)$. Note that this doesn't make any assuptions on the relationship between $f_T(x_i)$ and $f_S(x_i)$ for any individual instance $x_i$.  Thus, rather than transferring the overall shape of the source function, we can transfer this pairwise similarity information.  For example, this property would imply that if two neighborhoods have similar house prices, then they'll each need a similar number of taxi pickups. This can be a weaker transfer assumption because instead of requiring \textit{pointwise} similarity it assumes \textit{pairwise} similarity.

In this work we propose Pairwise Similarity Transfer (Figure \ref{fig:example-shift}), which assumes Pairwise Similarity holds and seeks to transfer this property from source to target.  This can be seen in Figure \ref{fig:example-shift} in how both functions divide the instance space into the same three subsets and, while the functions within these subsets are not identical, the first two are both relatively smooth (but the last is not).  As such, Pairwise Similarity Transfer doesn't fit into any of the previously proposed transfer learning settings.  This difference is key because it enables Pairwise Similarity Transfer based methods to outperform previous work when this modeling assumption is accurate.  Along these lines, we will show in our experiments that many real world data sets fall into this setting, such as the previously noted neighborhood housing price and taxi pickup example.

\noindent
\textbf{A Graph Based Mechanism For Transfer.} Our method leverages this transfer assumption by first constructing a ``Pairwise Similarity Graph'' which simultaneously measures the similarity of pairs of instances as a function of the source function predictions and potentially other aspects defined by the user such as spatial proximity (Section \ref{sec:our-method-spatial}) and guidance from a domain expert (Section \ref{sec:our-method-guidance}).

Given this graph, we construct a penalty term which regularizes the discrepancy of pairs of predictions made by the target function.  Since transfer is performed using a regularizer, this is a general purpose strategy that can be used with many learning algorithms. In our work we pair it with a nonparametric regression method and show it leads to better generalization performance than standard transfer learning methods on several spatial data sets which seem to display this pairwise similarity property.  We also experimentally explore other data sets which \textit{do not} display this property, and discuss when our method is appropriate.

Our contributions are:
\begin{itemize}[noitemsep]
\item We propose Pairwise Similarity Transfer, a new transfer learning setting  wherein the pairwise similarity of source function predictions are transferred to the target (Section \ref{sec:our-method}).
\item We propose Pairwise Similarity Regularization Transfer, a general purpose graph regularization transfer framework which uses an intuitive ``Pairwise Similarity Graph'' to transfer this information (Section \ref{sec:our-method}).
\item We show how the Pairwise Similarity Graph can be modified to incorporate spatial distance (Section \ref{sec:our-method-spatial}), user generated similarity constraints (Section \ref{sec:our-method-guidance}) and how to scale our method to large data sets using the Nystr{\"o}m method (Section \ref{sec:our-method-nystrom}).
\item Our theoretical result (Theorem \ref{thm:error}) show how the accuracy of the Pairwise Similarity transfer assumption impacts the generalization performance of our method (Section \ref{sec:theory}).
\item We present both positive and negative experimental results on synthetic and real data sets.  We also discuss when and why our methods perform better or worse than previous work (Section \ref{sec:experiments}).
\end{itemize}

The overview of our paper is as follows.  In Section \ref{sec:related-work} we discuss related work.  In Section \ref{sec:our-method} we introduce our method and compare and contrast it with previous work.  In Section \ref{sec:extensions} we present extensions to our method. In Section \ref{sec:theory} we present theoretical results about our method. In Section \ref{sec:experiments} we experimentally compare different versions of our method and previous work.  Finally we conclude in Section \ref{sec:conclusion} and discuss future work.

\section{Related Work}
\label{sec:related-work}

Transfer learning methods attempt to improve generalization performance using data from related source domains \cite{pan2010survey}.  Transfer learning algorithms vary in what relationships they assume hold between the source and the target, but most previous focuses on Covariate Shift and Hypothesis Transfer, which either implicitly or explicitly assume the corresponding predictive functions $f_S(x)$ and $f_T(x)$ are identical \cite{sugiyama2007covariate, sugiyama2008direct, gretton2009covariate, pan2011domain}  or almost identical \cite{daume2006domain, tommasi2010safety, gong2012geodesic, fernando2013unsupervised, kuzborskij2013stability}.  Exceptions to this include \cite{zhang2013domain, wang2014active, wang2014flexible}, which propose learning ``Location-Scale'' transformations to account for discrepancies between the domains.  

\subsection{Location-Scale Transfer.} Previous work proposed using ``Location-Scale'' transformations (compositions of scaling and translation functions) in order to adapt the source data to the target domain \cite{zhang2013domain, wang2014active, wang2014flexible}.  These methods first use what labeled target data is available to learn these transformations and then they augment the target data with the transformed source data.  All these methods assume the scaling and translation functions are relatively simple in order to prevent overfitting, and as we show these methods can perform poorly on real world data.

\subsection{Graph Regularization.} Laplacian Regularization is a popular framework for semisupervised learning  \cite{belkin2006manifold}.  These methods first construct a weighted graph $W$ over both the labeled and unlabeled training data where $W_{ij}$ is the edge weight between instances $i$ and $j$.  They then use its Laplacian, $D-W$, during the learning process to regularize the prediction discrepancy between pairs of instances $(f(x_i)-f(x_j))^2$, where $D$ is a diagonal matrix with entries equal to the row sums of $W$ \cite{von2007tutorial}.  While our regularization formulation is similar, our work differs in how the graph is constructed and what it represents.  Also, to our knowledge, using graph regularization for transferring pairwise similarity is novel.

\section{Our Method}
\label{sec:our-method}

\begin{algorithm}[t]
\begin{algorithmic}[1]
\REQUIRE
\STATEx\hspace{\algorithmicindent} $T = \{X_L,Y\}$: Labeled target data
\STATEx\hspace{\algorithmicindent} $T_U = \{X_U\}$: Unlabeled target data
\STATEx\hspace{\algorithmicindent} $f_S(x)$: Source function
\STATEx\hspace{\algorithmicindent} $k_1$: Source prediction function
\STATEx\hspace{\algorithmicindent} (Optional) $k_2$: Spatial kernel function
\ENSURE 
\STATEx\hspace{\algorithmicindent} $\hat{Y}$: Predictions on $X_L \cup X_U$
\STATEx

\FOR{all pairs $(x_i, x_j)$ in $X_L \cup X_U$}
\STATE $W_{ij} \leftarrow k_1(f_S(x_i) - f_S(x_j))$
\IF{$k_2$ provided}
\STATE $W_{ij} \leftarrow W_{ij} k_2(x_i - x_j)$ 
\ENDIF
\ENDFOR
\STATE $D \leftarrow $ Diagonal matrix of $W$'s row sums
\STATE $L \leftarrow D - W$
\STATE $f_T(x) \leftarrow $ solve equation \ref{eqn:region-guidance}
\STATE $\hat{Y} \leftarrow [ f_T(X), f_T(X_U)]$
\end{algorithmic}
\caption{Pairwise Similarity Regularization Transfer. }
\label{alg:ourMethod}
\end{algorithm}


We assume we are given a labeled target data set $T = (X_L,Y) = \{(x_1, y_1),...,(x_n, y_n)\}$, an unlabeled target data set $T_U = (X_U) = \{x_1,...,x_m\}$ and a source function $f_S(x)$ which was trained on a source data set $S$.  Importantly, we \textit{do not} assume the ground truth target function is identical or near identical to $f_S(x)$.  Rather, Our work assumes that, to some extent, the source and target have the Pairwise Similarity property:
\begin{defn}
A source and target domain with ground truth predictive functions $f_S(x)$ and $f_T(x)$ are \textbf{Pairwise Similarity} if, for much of the instance space, $f_S(x_i) \approx f_S(x_j)$ implies $f_T(x_i) \approx f_T(x_j)$.
\end{defn}

Note that this definition doesn't require this pairwise similarity to hold for the entire instance space.  

We propose modeling this pairwise information using a weighted graph.  Specifically, we treat each instance of the target domain as a node and for every pair of instances we compute a weight $W_{ij} = k(f_S(x_i) - f_S(x_j))$ where $k$ is a kernel, such as the Uniform or Gaussian kernels, and the weights are based on the \textit{source function predictions}.  These weights will be large when the source prediction discrepancy is small and vice versa.  Using these weights we use soft constraints to regularize the target function estimate $f_T(x)$ to enforce Pairwise Similarity.  In a manner analogous to Laplacian Regularization, we propose the following optimization problem:
\begin{align}
\label{eqn:region-guidance}
\min_{f_T(x)} & \sum_{x_i, y_i \in T} \mathcal{L}(f_T(x_i),y_i) +  \\
& \lambda \sum_{x_i \in X_L \cup X_U} \sum_{x_j \in X_L \cup X_U} W_{ij} (f_T(x_i) - f_T(x_j))^2 \nonumber \\
& W_{ij} = k(f_S(x_i) - f_S(x_j)) \ \ \ \forall x_i, x_j \in X_L \cup X_U \nonumber
\end{align}
where $\mathcal{L}$ is a loss function, $\lambda$ is a regularization parameter and for the moment we do not assume a specific form for $f_T(x)$.  The first term is the training error of the learned function on the labeled data, while the second is a graph based regularizer which penalizes deviations in the pairwise discrepancies between $f_S(x)$ and $f_T(x)$.  For example, if $W_{ij}$ is large (indicated $|f_S(x_i) - f_S(x_j|$ is small), then a large penalty will be incurred if $(f_T(x_i) - f_T(x_j))^2$ is large.  Conversely, if $W_{i, j} = 0$, then $f_T(x)$ can vary an arbitrary amount (with respect to these two instances).


Let $L = D - W$ be the graph Laplacian constructed from $W$, where $D$ is a diagonal matrix with entries equal to the row sums of $W$, and let $f_T(X)$ be the vector of predictions made by $f_T(x)$ on the entire data set $X = X_L \cup X_U$.  Using this notation, the latter term of equation \ref{eqn:region-guidance} can be more concisely written in matrix form  \cite{belkin2006manifold}:
\begin{align}
f_T(X)' L f_T(X) = \sum_{x_i \in X} \sum_{x_j \in X} W_{ij} (f_T(x_i) - f_T(x_j))^2
\label{eqn:reg-matrix-form}
\end{align}
While this regularization framework does not assume a specific form for $f_T(x)$, we experiment with the Nadaraya-Watson estimator (NW) \cite{hastie2005elements}, a nonparametric regression estimator.  The first step of NW is to construct a similarity matrix $W_{NW}$, a $m \times n$ matrix which measures the similarity between the $n$ labeled training instances and the $m$ instances to predict.  Next, letting $D_{NW}$ be a diagonal matrix constructed from the row sums of $W_{NW}$, the smoothing matrix $M_{NW}=D^{-1}_{NW} W_{NW}$ is calculated.  Finally, letting $f_{NW}$ be the predictions, the NW solution is $f_{NW} = M_{NW} Y$.  In order to incorporate our regularizer, note that NW implicitly solves the following optimization problem:
\begin{align}
\min_{f_{NW}} ||f_{NW} - M_{NW} Y||^2
\end{align}
Thus, in order to incorporate Pairwise Similarity Regularization we add the graph regularizer from equation \ref{eqn:reg-matrix-form} to get:
\begin{align}
\label{eqn:region-guidance-nw}
\min_{f_T(X)} || f_T(X) - M_{NW} Y||^2 + \lambda f_T(X)' L f_T(X)
\end{align}
The first term is simply the objective of the Nadaraya-Watson estimator, while the second is our graph based regularizer.

This formulation has the closed form solution $f_T(X) = (I + \lambda L)^{-1} M Y$, which we derive in the supplementary materials.  Also, while this formulation only makes predictions on the given data $X_L \cup X_U$, out of sample extensions can be made by using the data set $(X, f_T(X)) \cup (X_U, f_T(X_U))$ to train any standard supervised learning model.

\subsection{Comparison With Previous Transfer Learning Assumptions. }
Existing transfer learning formulations such as the first three in Figure \ref{fig:example-shift} make strong assumptions on the similarity of the source and target functions. Specifically, $\forall x: || f_S(x) - f_T(x) || \le \epsilon$ (Location-Scale makes this assumption after the learned transformations are applied to the source data).  Pairwise Similarity can be seen as a \textit{weaker} assumption than these previous transfer learning settings for several reasons. First, it assumes a relationship between pairs of predictions, without requiring $f_T(x)$ to $f_S(x)$ to have similar shapes.  Second, it makes assumptions on subsets of the instance space where pairs of similar predictions are made in the source, while it makes no assumptions on the remainder of the instance space.  Finally, since it is implemented as a regularizer, the impact of the assumption can be controlled by tuning $\lambda$.

\section{Extensions}
\label{sec:extensions}
Here we outline three extensions to our work: incorporating spatial continuity, adding human guidance to the construction of the graph and scaling our method to large data sets.

\subsection{Enforcing Spatial Continuity. }
\label{sec:our-method-spatial}
The previous formulation only considers source prediction discrepancies when constructing $W$, but in many settings also considering \textit{spatial} discrepancies between pairs of instances can be useful. 
Returning to the taxi pickup example, 
we might want to disentangle two neighborhoods which have similar housing prices but lie on different sides of the city.   To do this, we modify the graph to include a spatial component as well:
\begin{align}
W_{ij} = k_1(f_S(x_i) - f_S(x_j)) k_2(x_i - x_j)
\end{align}
where $k_1$ is the kernel over the source prediction discrepancies and $k_2$ is the kernel over the spatial discrepancies.  Using this graph, $W_{ij}$ will only be large if the instances have similar predictions in the source domain and are spatially close.

\subsection{Adding User Guidance. }
\label{sec:our-method-guidance}

\begin{algorithm}[t]
\begin{algorithmic}[1]
\REQUIRE
\STATEx\hspace{\algorithmicindent} $W$: Pairwise Similarity Graph
\STATEx\hspace{\algorithmicindent} $S$: Set of ``similar'' pairs
\STATEx\hspace{\algorithmicindent} $D$: Set of ``dissimilar'' pairs
\ENSURE 
\STATEx\hspace{\algorithmicindent} $\tilde{W}$: Graph with user guidance
\STATEx

\STATE $\tilde{W} \leftarrow W$
\FORALL{$(x_1, x_2) \in S$}
\STATE $\tilde{W_{ij}} \leftarrow 1$
\ENDFOR
\FORALL{$(x_1, x_2) \in D$}
\STATE $\tilde{W_{ij}} \leftarrow 0$
\ENDFOR
\end{algorithmic}
\caption{Modifying the Smoothness Graph using pairwise guidance. }
\label{alg:guidance}
\end{algorithm}

One advantage of using a graph based regularizer is it can be easily modified to incorporate pairwise constraints generated by the user or from side information.  This information can improve the performance of our method by replacing entries in $W$ which $f_S(x)$ gets ``wrong.''  Clearly the ideal guidance of this form would be to replace entries in $W$ using an ``Oracle'' which generates weights using the ground truth target labels, but in practice, users are unlikely to be able to generate accurate estimates of the delta $y_i - y_j$, so we instead propose the use of \textit{binary} guidance, where each modified entry in $W$ is set to $0$ or $1$ (see Algorithm \ref{alg:guidance}).  For example, if the user knows $y_i$ and $y_j$ should be similar, then they could set $W_{ij} = 1$.  Alternatively, the user could set $W_{ij} = 0$ if they do not think they're related.

While this information is less precise, it can be more easily generated by users.  We discuss the theoretical aspects of this guidance in Section \ref{sec:theory} and experimentally test the impact of it in Section \ref{sec:experiments}.

\subsection{Scaling to Large Data Sets. }
\label{sec:our-method-nystrom}

\begin{algorithm}[t]
\begin{algorithmic}[1]
\REQUIRE
\STATEx\hspace{\algorithmicindent} $W$: Pairwise Similarity Graph
\ENSURE 
\STATEx\hspace{\algorithmicindent} $V$: Approximate solution to $(I + \lambda L)^{-1}$
\STATEx
\STATE $D \leftarrow$ Diagonal matrix of $W$'s row sums
\STATE $C \leftarrow$ A randomly sampled submatrix of $p$ of $W$'s columns
\STATE $B \leftarrow$ The upper $p \times p$ block of $C$
\STATE $M \leftarrow$ $(I + \lambda D)^{-1}$
\STATE $V \leftarrow M(I - C(-B + C'MC)^{-1}C'M)$
\end{algorithmic}
\caption{Matrix inverse approximation using the Nystr{\"o}m approximation.}
\label{alg:nystrom}
\end{algorithm}

The closed form solution to our method requires inverting a large matrix (or solving a large system of linear equations), which runs in time cubic with respect to the number of instances in $(X_L \cup X_U)$ \cite{golub2012matrix}.  This can be too computationally expensive to run on large data sets, so we scale our method by approximating $W$ using the Nystr{\"o}m method, a column sampling method for matrix sketching \cite{fowlkes2004spectral}.  

Using Nystr{\"o}m, we approximate $W$ as $C B^\dagger C'$ where $C$ are the $p$ sampled columns, $B$ is the upper $p\times p$ block of $C$ and $B^\dagger$ denotes the pseudoinverse of $B$.  Using the Woodbury matrix identity \cite{horn2012matrix}, and letting $M = (I + \lambda D)^{-1}$, we have $(I + \lambda(D - C B^\dagger C'))^{-1} = M(I - C(-B + C'MC)^{-1}C'M)$ (see Algorithm \ref{alg:nystrom}).  The latter term requires inverting a diagonal matrix and a dense matrix of size $p \times p$ matrix where $p << n$, both of which can be done significantly faster.  We found that setting $p$ to a tiny fraction of $n$ had a minor impact on generalization performance for most data sets while dramatically improving running time.

\section{An Error Bound}
\label{sec:theory}

\newcommand{\xxi}{x_i}
\newcommand{\xxj}{x_j}
\newcommand{\ft}{f_{T}^*}
\newcommand{\fs}{f_{S}^*}
\newcommand{\lt}{L_{T}}
\newcommand{\ls}{L_{S}}
\newcommand{\fltHat}{\hat{f}_{L_T}}
\newcommand{\flsHat}{\hat{f}_{L_S}}
\newcommand{\ct}{C_{T}}
\newcommand{\cs}{C_{S}}
\newcommand{\y}{Y}
\newcommand{\yHat}{\hat{f}_{NW}(X)}
\newcommand{\yHatT}{\hat{f}_T(X)}
\newcommand{\yHatS}{\hat{f}_S(X)}
\newcommand{\dy}{\delta_Y}
\newcommand{\dc}{\Delta_C}
\newcommand{\dl}{\Delta_L}

In this section we provide an error bound for our algorithm.  This bound is useful because it provides theoretical justification for our algorithm and shows when our algorithm will be effective.  The proof is included in the supplementary materials.

First, some notation:
\begin{itemize}[noitemsep]
\item $\ft(x), \fs(x)$: Ground truth target and source functions
\item $\lt$: Laplacian created using $\ft(x)$.  i.e. $\lt = D - W$ where $W_{ij} = k(\ft(\xxi) - \ft(\xxj))$
\item $\ls$: Laplacian created using $\fs(x)$.  i.e. $\ls = D-W$ where $W_{ij} = k(\fs(\xxi) - \fs(\xxj))$
\item $\yHat, \yHatT, \yHatS$: Vectors of predictions on the test data using Nadaraya Watson, our method with $\lt$, and our method with $\ls$ respectively.
\end{itemize}

Our goal is to bound the error $||\ft(X) - \yHatS||$, the error of our method when using the source Laplacian $\ls$.

\begin{Theorem}[Error Bound]
The prediction error of our method is bounded by:
\label{thm:error}
\begin{align*}
||\yHatS - \ft(X)|| \leq \mathcal{O}(||\yHatT - \ft(X)|| + \\
\lambda ||\ls - \lt||(1 + ||\ft(X) - \yHat||))
\end{align*}
\end{Theorem} 

The first term on the right hand side is the error of our method when using the ``correct'' Laplacian $\lt$.  This ``approximation error'' is a property of $T$ and the current hyperparameters \cite{bottou2010large}.  
 
The second term depends on two quantities: the error of the NW estimate $\yHat$ and the difference between the source and target Laplacians.  Ideally, we would simply use $\lt$, but because $\ft(x)$ is unavailable it cannot be used in practice.  However, this term will be small when the source and target prediction graphs are very similar, showing our method can be successful if an appropriate source is used.  While this term is minimized by setting $\lambda$ to $0$, doing so would make $\yHatT = \yHat$, which would cause the first term to grow.

\subsection{Adding Constraints. }

Suppose the user adds guidance to $\ls$, as suggested in section \ref{sec:our-method-guidance}, to produce $\tilde{L}_S$.  In this case, the previous bound becomes a function of $||\tilde{L}_S - \lt||$.  Thus, adding guidance is theoretically  motivated as long as the new Laplacian better approximates $\lt$.  This has two consequences.  First, it shows that adding constraints can improve the accuracy of this method.  Second, it shows that even adding \textit{noisy} guidance can be helpful as long as it makes it more similar to $\lt$.

\section{Experiments}
\label{sec:experiments}
In our experiments we attempt to answer the following questions regarding the accuracy and scalability of our method, as well as the impact of adding simulated human guidance:
\begin{itemize}[noitemsep]
\item How well does our method perform compared with previous transfer learning work and standard baseline methods (Tables \ref{tab:positive-all-methods} and \ref{tab:negative-all-methods})?
\item What is the impact of including spatial continuity when constructing the Pairwise Similarity Graph (Table \ref{tab:guidance-nystrom})?
\item How does using the Nystr{\"o}m method effect the performance of our method (Figure \ref{tab:guidance-nystrom})?
\item Does incorporating domain knowledge in the form of pairwise constraints when constructing the graph improve the accuracy of our method (Table \ref{tab:guidance-nystrom})?
\end{itemize}

\begin{table*}[t]
\footnotesize
\centering
\begin{tabular}{| p{3.8cm} | p{13cm} |}
\hline
Method & Description 
\\ \hline
\textbf{Target Only} &Training using only the target data.
\\ \hline
\textbf{Semisupervised \cite{zhou2004learning}} & Semisupervised regression using the ``Learning with Local and Global Consistency'' method.
\\ \hline
\textbf{Stacked \cite{pan2010survey}} & An Hypothesis Transfer method, where predictions made by learned source and target functions are used as features for a linear function.  This method performs best when the source and target functions are close to identical.  While simple, this method can perform very well in practice.
\\ \hline
\textbf{Offset \cite{wang2014active}} & A state-of-the-art transfer learning method which account for differences between the source and target by estimating a translation transformation between the source and target domains.
\\ \hline
\textbf{Location-Scale \cite{wang2014flexible}} & A state-of-the-art transfer learning method which estimates a location-scale transformation to map the source data to the target task.
\\ \hline
\end{tabular}
\caption{Competing methods we experimented with. \label{tab:competing-methods}}
\end{table*}

The competing methods we used are described in Table \ref{tab:competing-methods}.  The variations of our method we used are:

\noindent
\textbf{PSRT:} The Pairwise Similarity Regularization Transfer method we proposed (Section \ref{sec:our-method}).

\noindent
\textbf{PSRT+SC:} The Pairwise Similarity Regularization Transfer method we proposed, including spatial continuity (Section \ref{sec:our-method-spatial}).

\noindent
\textbf{PSRT: $\%N$ Nystr{\"o}m:} Our method where we used the Nystr{\"o}m method to accelerate our solver (Section \ref{sec:our-method-nystrom}), where $N$ is the percentage of columns sampled.

\noindent
\textbf{PSRT: $\%N$ Guidance:} Our method with guidance (Section \ref{sec:our-method-guidance}), where some percentage of the Similarity Graph entries were replaced with $1$ if the ground truth target labels were similar and $0$ otherwise.  Here, we defined ``similar'' to be if the absolute difference in labels was within the bottom tenth percentile of all pairwise differences.  $N$ is the percentage of the possible pairs which were sampled for guidance.

\textbf{Because the data sets we use do not suffer from ``Covariate Shift,'' we did not run experiments using Covariate Shift and Domain Adaptation methods such as those in \cite{sugiyama2007covariate, sugiyama2008direct, gretton2009covariate, pan2011domain}.}  However, these methods could be used a preprocessing step for our method when using data with Covariate Shift.  For all methods all parameters were tuned on a validation set and results are the average on 30 train/test splits.  Values in parentheses are $95\%$ confidence intervals.  
\textbf{If this paper is accepted then all code and data will be made available online.}

\begin{figure}[h]
  \centering
   \begin{subfigure}[b]{1\columnwidth}
\includegraphics[width=1\columnwidth]{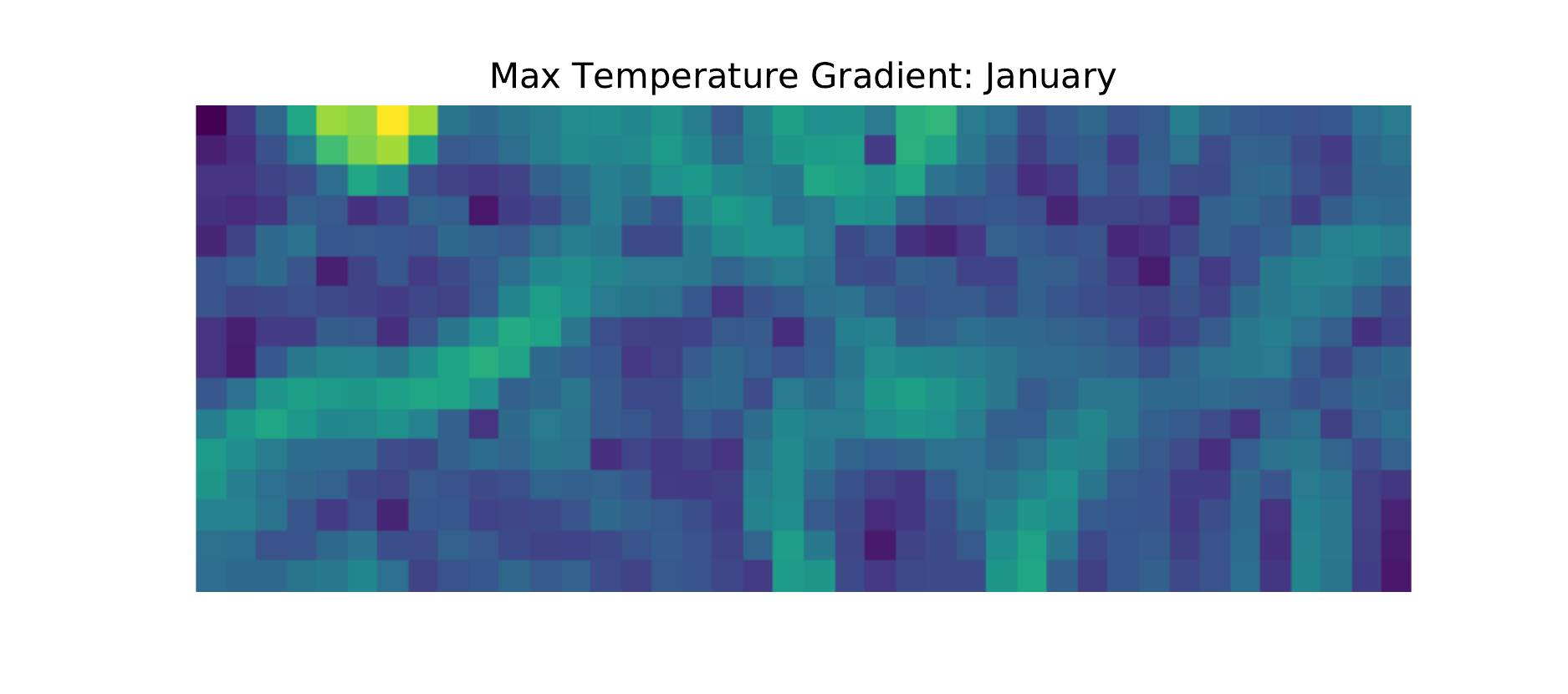}
  \caption{Gradient magnitude of high recorded temperature in January.} 
  \label{fig:climate-january-gradient}
\end{subfigure}
  \centering
    \begin{subfigure}[b]{1\columnwidth}
\includegraphics[width=1\columnwidth]{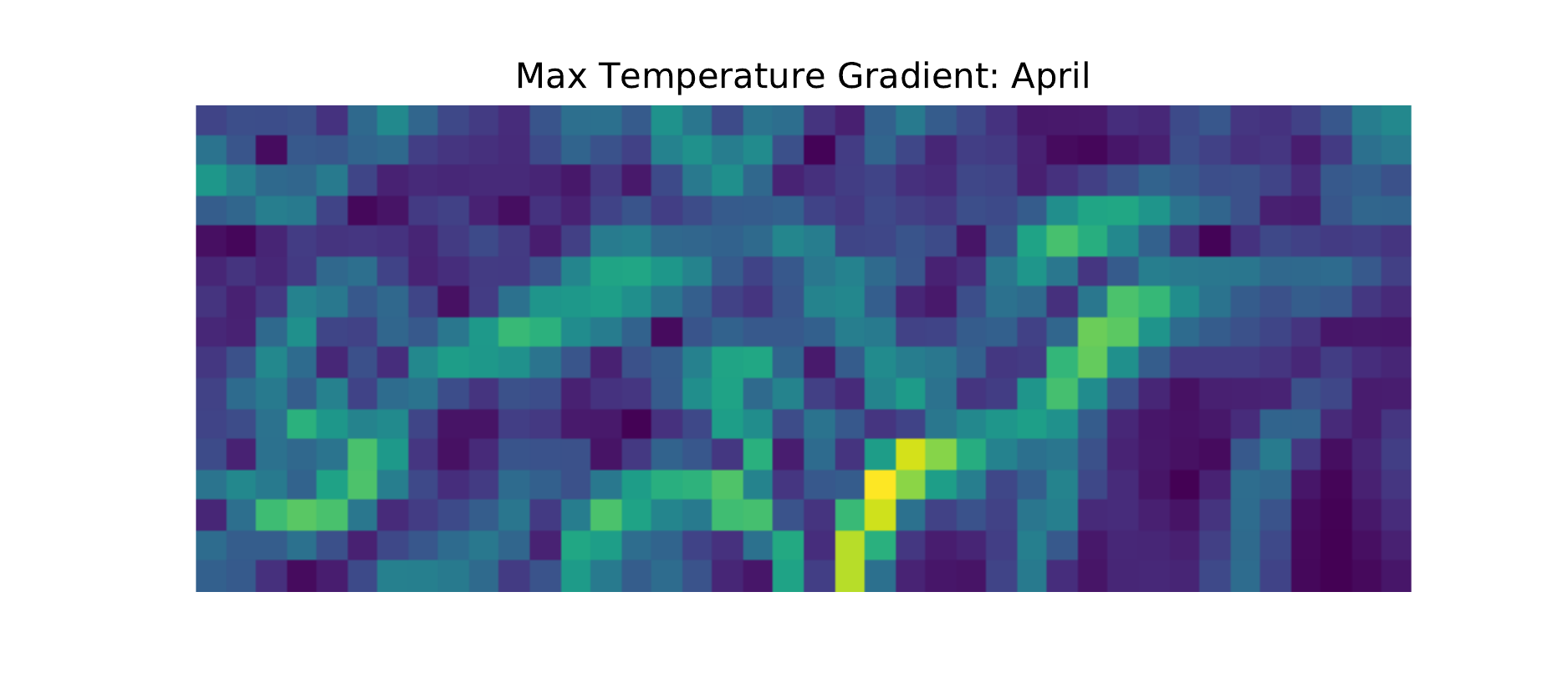}
  \caption{Gradient magnitude of high recorded temperature in April.} 
  \label{fig:climate-april-gradient}
  \end{subfigure}
  
  \caption{Gradient magnitudes of temperature data.  Note the shared locations where the gradient magnitude is small, implying the predictions between pairs of instances within these regions will be similar.}
  \label{fig:climate-gradient}
\end{figure}

\begin{figure}[h]
  \centering
     \begin{subfigure}[b]{.4\columnwidth}
\includegraphics[width=\columnwidth, trim={6cm 1cm 6cm 0cm}, clip]{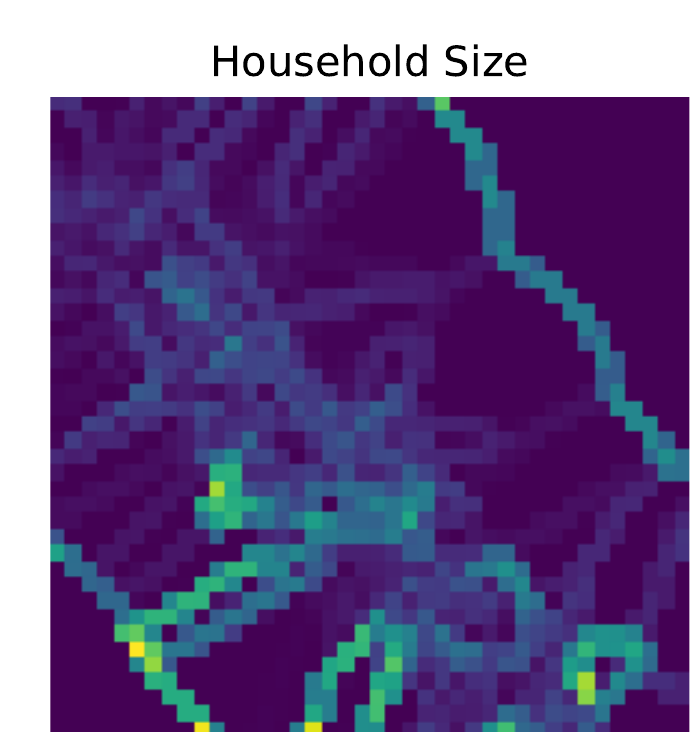}
  \label{fig:census-household-gradient}
\end{subfigure}
     \begin{subfigure}[b]{.4\columnwidth}
\includegraphics[width=\columnwidth, trim={6cm 1cm 6cm 0cm}, clip]{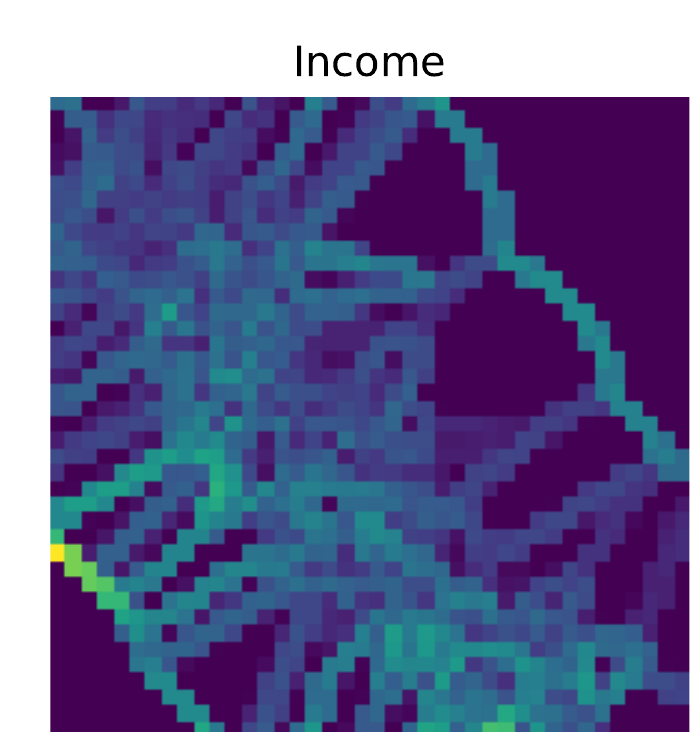}
  \label{fig:census-income-gradient}
 \end{subfigure}
\caption{Gradient magnitudes of census data.  Note the shared locations where the gradient magnitude is small.}
   \label{fig:census-gradient}
\end{figure}

\begin{figure}[h]
  \centering
     \begin{subfigure}[b]{.4\columnwidth}
\includegraphics[width=\columnwidth, trim={6cm 1cm 6cm 0cm}, clip]{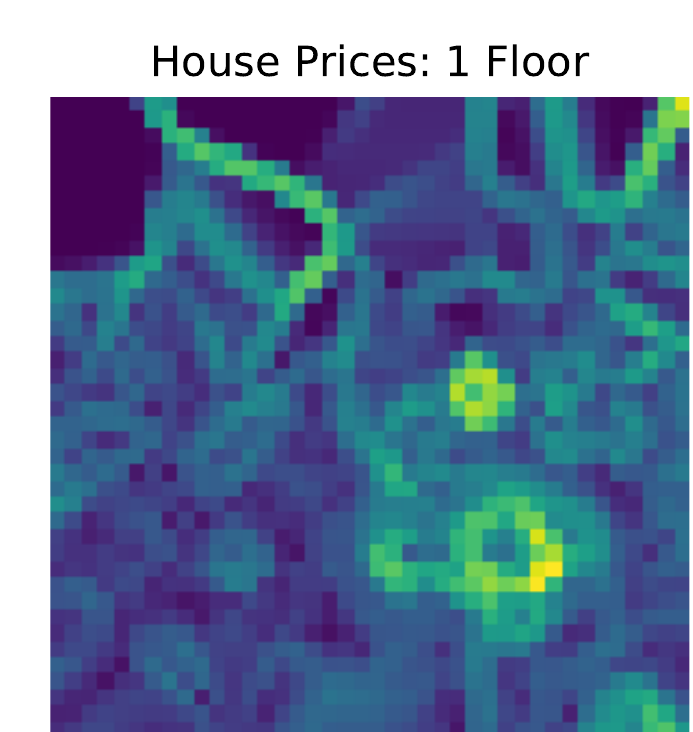}
  \label{fig:kc-1floor-gradient}
\end{subfigure}
     \begin{subfigure}[b]{.4\columnwidth}
\includegraphics[width=\columnwidth, trim={6cm 1cm 6cm 0cm}, clip]{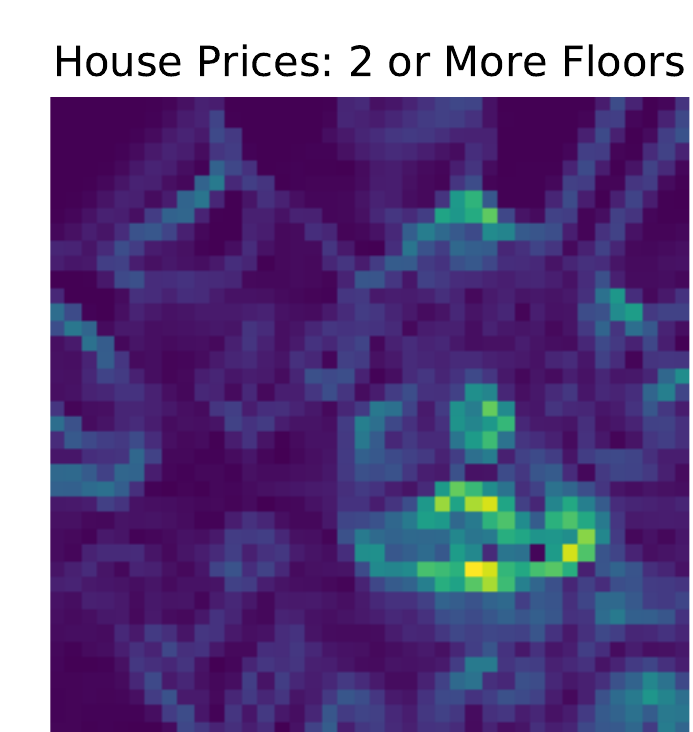}
  \label{fig:kc-2floor-gradient}
\end{subfigure}

\caption{Gradient magnitudes of King County housing data.  While our method performed well on this data set, the Offset method performed better, suggesting that while the pairwise similarity information was useful, the source-to-target transformation was simple enough to use simpler transfer learning methods.}
  \label{fig:kc-gradient}
\end{figure}

\textbf{Data:} We used 7 regression data sets: four spatial data sets which we found Pairwise Similarity Transfer performed well on, and three data sets where Pairwise Similarity Transfer did not perform well.  This is to be expected as the Pairwise Similarity transfer assumption, like any transfer learning assumption, need not always hold. We included both positive and negative experiments to better understand the benefits and limitations of Pairwise Similarity Transfer.  The data sets are described in Tables \ref{tab:data-positive} and \ref{tab:data-negative}.  To better understand the performance of our method we also included heat maps of the gradient magnitudes of the predictive functions in Figures \ref{fig:climate-gradient}, \ref{fig:census-gradient} and \ref{fig:kc-gradient}.  These plots were generating by estimating the gradients of the predictive functions through the instance space and visualizing the norms of the gradients, where lighter values correspond to a lower magnitude.  e.g. Darker areas indicate less variation of the function.

\begin{table*}[t]
\footnotesize
\centering
\begin{tabular}{| p{3.8cm} | p{13cm} |}
\hline
Data Set & Description 
\\ \hline
\textbf{Synthetic Piecewise} & A synthetic piecewise constant data set with Gaussian noise.  The source and target data were different piecewise constant functions, but share the same set of discontinuities.
\\ \hline
\textbf{Census \cite{USCensus}} & Predicting the mean household size of zip codes in southern California as recorded from the 2010 US census.  Source is mean income of households, as recorded in the 2014 American Community Survey.
\\ \hline
\textbf{Temperature \cite{climate}} & Predicting the mean, high temperature in April 2016 across the southern United States.  Source is mean, high temperature from January 2016.
\\ \hline
\textbf{Taxi+Housing \cite{zillow, taxi}} & Predicting the average number of taxi pickups at various regions of San Francisco, California, between 5am and 12pm over thirty days.  Source is average home price of neighborhood.
\\ \hline
\end{tabular}
\caption{Data sets for which Pairwise Similarity Transfer perform welled. All of the real data sets were complicated spatial prediction problems with nuanced differences between the source and target domains.\label{tab:data-positive}}
\end{table*}

\begin{table*}[t]
\footnotesize
\centering
\begin{tabular}{| p{3.8cm} | p{13cm} |}
\hline
Data Set & Description 
\\ \hline
\textbf{Bike Sharing \cite{fanaee2014event, Lichman:2013}} & Bike rental prediction in 2011 and 2012 as a function of weather.  We used the 2011 data as the source and the 2012 data as the target.
\\ \hline
\textbf{Boston Housing (BH) \cite{harrison1978hedonic, Lichman:2013}} & House price prediction in Boston as a function of number of rooms in the house.  Bottom quartile of LSTAT (percentage of lower status of the population) for target, second quartile for source.
\\ \hline
\textbf{King County Housing (KC Housing) \cite{kcHousing}} & Predicting housing prices in King County, Washington as a function of location.  We split houses into source and target data sets by the number of floors of each house.
\\ \hline
\end{tabular}
\caption{Data sets for which Pairwise Similarity Transfer did not perform as well as competing methods. The source and target domains were much similar, so simpler transfer learning methods tended to perform best.  \label{tab:data-negative}}
\end{table*}

\begin{figure}[t]
\centering
  \begin{subfigure}[b]{1\columnwidth}
\includegraphics[width=1\textwidth]{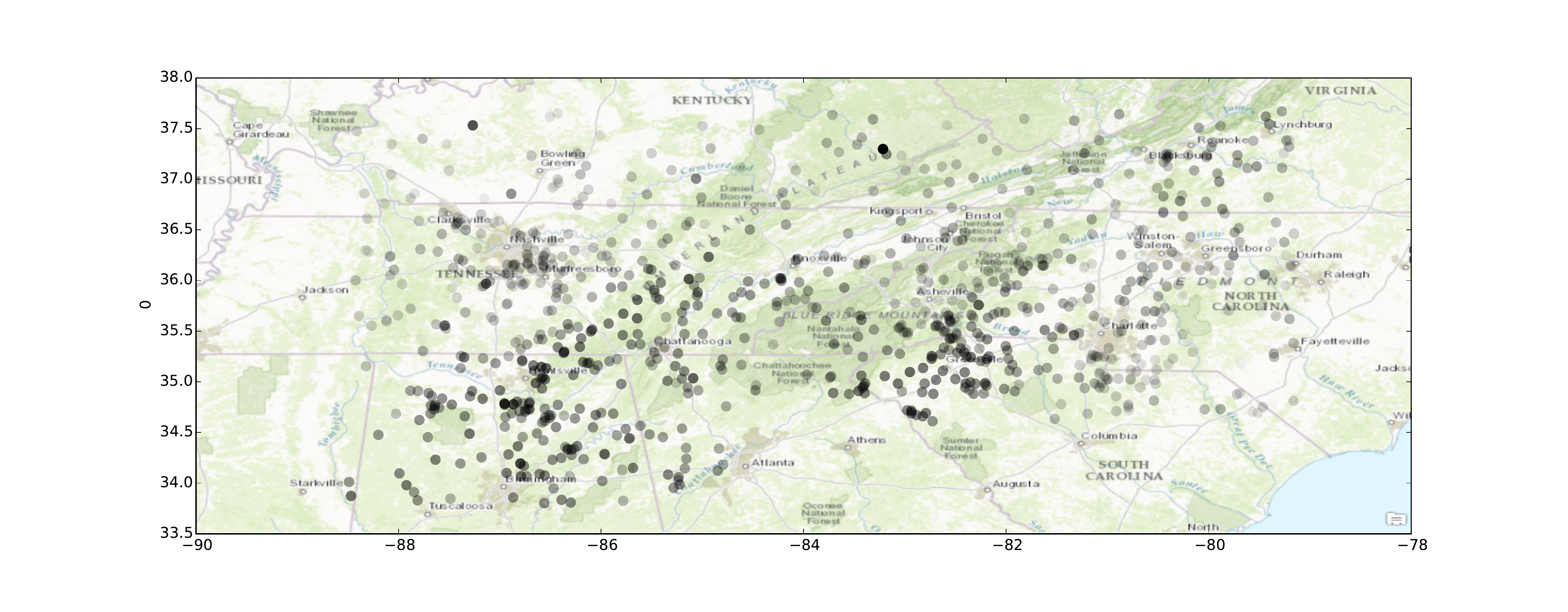}
  \caption{High recorded temperature in January.} 
  \label{fig:climate-january}
\end{subfigure}

\centering
\begin{subfigure}[b]{1\columnwidth}
\includegraphics[width=1\textwidth]{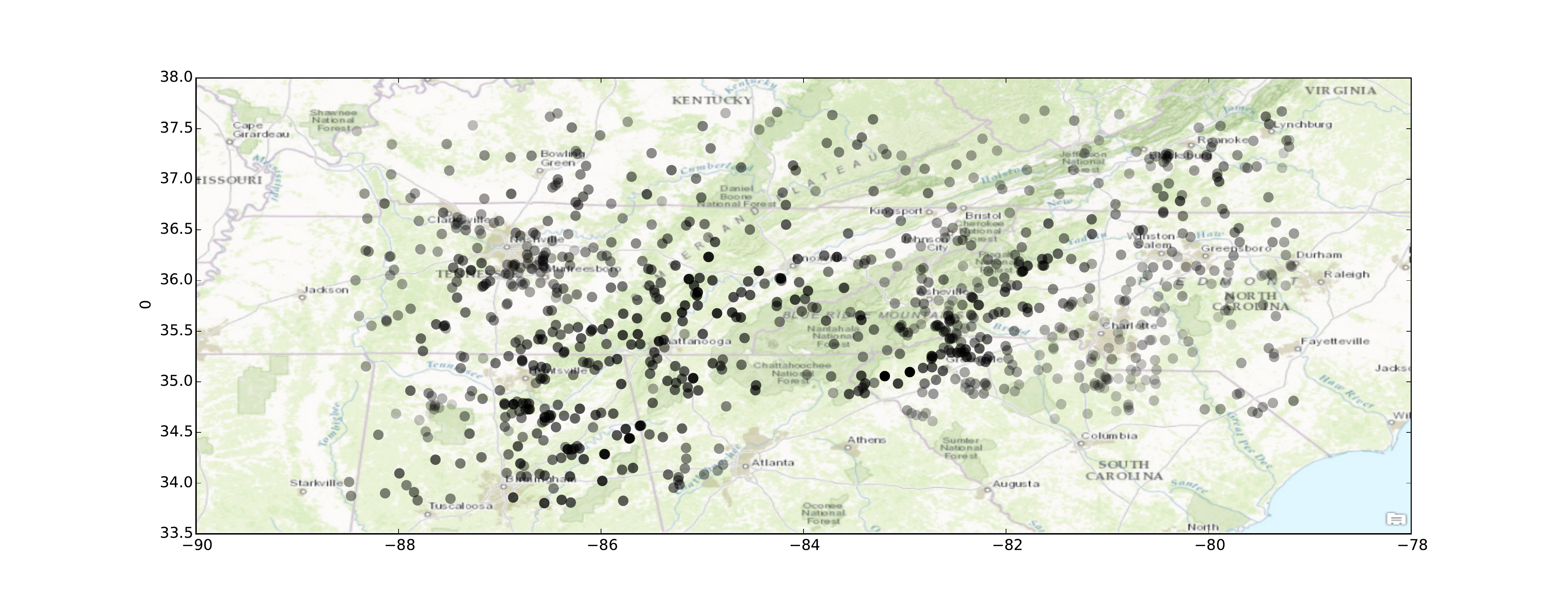}
  \caption{High recorded temperature in April.} 
  \label{fig:climate-april}
  \end{subfigure}
  
  \caption{High recorded temperatures in the southern United States.  Darker values indicate lower temperature.  Note that data is smooth in the majority of sites but quickly shifts in a handful of regions that are shared between the two data sets.}
\end{figure}

\begin{table*}[h]
\resizebox{\textwidth}{!}{
\centering
\begin{tabular}{| m{2cm} | r | r | r| r | r | r | r | r |}
\hline
 & PSTR+SC & PSTR& Location-Scale & Semisupervised & Target Only & Stacked & Offset\\ \hline 
Piecewise &  $\textbf{0.055 (0.01)}$ & $0.062 (0.01)$ & $0.653 (0.08)$ & $0.524 (0.06)$ & $0.108 (0.03)$ & $0.147 (0.02)$ & $0.178 (0.04)$ \\ \hline
Census &  $0.200 (0.02)$ & $\textbf{0.190 (0.02)}$ & $0.375 (0.05)$ & $0.215 (0.03)$ & $0.206 (0.03)$ & $0.209 (0.03)$ & $0.206 (0.03)$ \\ \hline
Temperature &  $\textbf{0.127 (0.01)}$ & $0.129 (0.01)$ & $0.643 (0.08)$ & $0.280 (0.02)$ & $0.148 (0.01)$ & $0.165 (0.01)$ & $0.157 (0.01)$ \\ \hline
Taxi+Housing &  $\textbf{0.254 (0.02)}$ & $0.272 (0.01)$ & $0.710 (0.06)$ & $0.426 (0.04)$ & $0.303 (0.03)$ & $0.293 (0.02)$ & $0.287 (0.02)$ \\ \hline
\end{tabular}
}
\caption{Mean squared error of ours and competing methods on spatial data sets which seemed to exhibit Pairwise Similarity, training with 20 labeled target instances.  Values in parentheses are $95\%$ confidence intervals.  Our method performed much better than previous work.  Also, the other transfer learning methods generally displayed negative transfer, while ours always outperforms using just the target data.
\label{tab:positive-all-methods}}
\end{table*}

\begin{table*}[h]
\resizebox{\textwidth}{!}{
\centering
\begin{tabular}{| m{2cm} | r | r | r| r | r | r | r | r |}
\hline
 & PSTR+SC & PSTR & Location-Scale & Semisupervised & Target Only & Stacked & Offset\\ \hline 
Bike Sharing &  $0.072 (0.01)$ & $0.073 (0.01)$ & $0.436 (0.12)$ & $0.167 (0.02)$ & $0.082 (0.01)$ & $0.080 (0.01)$ & $\textbf{0.066 (0.01)}$ \\ \hline
BH &  $0.303 (0.05)$ & $0.309 (0.05)$ & $0.532 (0.12)$ & $0.899 (0.18)$ & $0.318 (0.05)$ & $0.387 (0.06)$ & $\textbf{0.280 (0.05)}$ \\ \hline
KC Housing &  $0.567 (0.06)$ & $0.565 (0.06)$ & $0.525 (0.05)$ & $0.761 (0.06)$ & $0.659 (0.06)$ & $0.565 (0.05)$ & $\textbf{0.497 (0.05)}$ \\ \hline
\end{tabular}
}
\caption{Mean squared error of ours and competing methods on data sets which did not seem to exhibit Pairwise Similarity, training with 20 labeled target instances.  Values in parentheses are $95\%$ confidence intervals.  On these data sets Offset performed best, but our method still outperformed other forms of transfer learning and using just the target data.
\label{tab:negative-all-methods}}
\end{table*}

\begin{table*}[h]
\resizebox{\textwidth}{!}{
\centering
\begin{tabular}{| m{2cm} | r | r | r | | r | r |}
\hline
 & PSTR+SC & PSTR: 10\% Guidance & PSTR: 20\% Guidance & PSTR: 10\% Nystrom & PSTR: 20\% Nystrom\\ \hline 
Piecewise &  $0.062 (0.01)$ & $0.063 (0.01)$ & $\textbf{0.058(0.01)}$ & $0.085 (0.01)$ & $0.085 (0.01)$ \\ \hline
Census &  $\textbf{0.190 (0.02)}$ & $0.193 (0.02)$ & $0.197 (0.02)$ & $0.193 (0.02)$ & $0.196 (0.02)$ \\ \hline
Temperature &  $0.129 (0.01)$ & $0.122 (0.01)$ & $\textbf{0.117 (0.01)}$ & $0.129 (0.01)$ & $0.129 (0.01)$ \\ \hline
Taxi+Housing &  $0.272 (0.01)$ & $0.219 (0.02)$ & $\textbf{0.212 (0.02)}$ & $0.272 (0.01)$ & $0.271 (0.01)$ \\ \hline
\end{tabular}
}
\caption{Mean squared error of ours method using simulated guidance and the Nystr{\"o}m approximation.  Values in parentheses are $95\%$ confidence intervals.  Incorporating guidance generally led to better generalization performance.  Using Nystr{\"o}m led to some drops in performance, but these drops were relatively small.
\label{tab:guidance-nystrom}}
\end{table*}

\textbf{All Methods Experiments:} Table \ref{tab:positive-all-methods} shows the performance of these methods on real and synthetic data sets.  For these data sets our method performed best.  The positive performance of our method seems to be due to Pairwise Similarity holding for these data sets.  For example, the climate data is shown in  Figures \ref{fig:climate-january}, \ref{fig:climate-april}, and the magnitude of the gradients are shown in Figure \ref{fig:climate-gradient}.  From these figures we see the two data sets seem to share subsets of the instance space where the functions are relatively smooth.  In particular, looking at the gradient information shows shared areas of near-zero gradient magnitude.  The other transfer methods likely performed worse due to their transfer assumptions being violated.

Table \ref{tab:negative-all-methods} shows the performance of these methods on a different set of data sets.  While our method generally performed better than Location-Scale, Target Only and Stacked,  Offset performed best overall.  This is likely due to the transformation between the source and target being smooth.  Looking at the gradient information in Figure \ref{fig:kc-gradient} gives us some insight into this.  While there is some alignment in where the functions do not rapidly vary, it is to a much lesser degree than in Figures \ref{fig:climate-gradient} and \ref{fig:census-gradient}.  This matches the empirical performance of our method, which was good, but not as strong as Offset.

It's worth noting how the ``worst case'' performance of our method and Offset differed.  While Offset displays \textbf{negative transfer} on most of the first four data sets (performing worse than just using the target data), our method \textbf{always} performed better than Target Only.  This is important because it suggests our method is more robust to negative transfer than Offset.

Another key insight is that all the real data sets with positive results were spatial data sets, indicating Pairwise Similarity is a property that can likely hold for spatial data.

\textbf{Spatial Continuity:} Tables \ref{tab:positive-all-methods} and \ref{tab:negative-all-methods} shows the performance of our method with and without the incorporation of spatial continuity when construction the Pairwise Similarity Graph.  These results show including it can have a dramatic improvement on performance, but did not universally help.  Rather, the utility of spatial continuity seems to be a property of data set, but using it seems wise in practice. 


\textbf{Nystr{\"o}m Approximation:}  Table \ref{tab:guidance-nystrom} shows the performance of our method when using the Nystr{\"o}m method to approximate the source prediction graph.  These results show that for most of our data sets sampling had a negligible impact on performance, even when only a small fraction of the Pairwise Similarity Graph's columns were used.  The exception to this is the Synthetic Piecewise data set, which displayed erratic performance.  We suspect this is due to the Pairwise Similarity Graph not being low rank.

Overall, these results demonstrate that our method can be scaled by using the Nystr{\"o}m approximation without significant losses in accuracy.

\textbf{Binary Guidance:} Table \ref{tab:guidance-nystrom} shows the performance of our method when incorporating varying amounts of binary guidance (see Section \ref{sec:our-method-guidance}).  These results show our approach of encoding guidance can dramatically improve the performance of our method.  This is an important result because it shows even the relatively noisy guidance that humans could provide can aid our algorithm.  Additionally, this shows that pairwise similarity guidance can be an effective mechanism for improving transfer learning.

\section{Conclusion}
\label{sec:conclusion}

We proposed Pairwise Similarity Transfer, a new transfer learning setting where the similarity between pairs of source predictions are transferred, and we empirically showed this assumption leads to positive results for a variety of complicated spatial transfer learning problems.  We modeled Similarity Transfer by creating a graph from the source data where each node is an instance and the edge weights are a function of the source prediction discrepancies. Modeling this form of transfer through a graph has several benefits. First, it is easy to extend the method to include spatial discrepancy.  Second, we can use the Nystr{\"o}m Approximation to scale the method to larger data sets.  Finally, this allows a novel method of encoding pairwise user guidance.  We showed that our method has strong theoretical justification by bounding the prediction error. In the future, we will explore whether Pairwise Similarity can occur in non spatial data sets.  In particular, we suspect it will hold in some time series data.

\bibliography{bib}

\end{document}